\documentclass[11pt,a4paper]{article}
\usepackage[hyperref]{emnlp2020}
\usepackage{times}
\usepackage{latexsym}
\renewcommand{\UrlFont}{\ttfamily\small}
\aclfinalcopy
\def\aclpaperid{148}
\usepackage{microtype}

\usepackage{verbatim}
\usepackage{amsmath}
\usepackage{bm}
\usepackage{booktabs}
\usepackage{subcaption}
\usepackage{float}
\usepackage{tipa}
\usepackage{comment}
\usepackage{graphicx}
\usepackage{xspace}
\usepackage{multirow}
\usepackage[normalem]{ulem}
\useunder{\uline}{\ul}{}
\usepackage[inline, shortlabels]{enumitem}
\usepackage{xcolor}
\definecolor{green}{HTML}{268B07}
\definecolor{blue}{HTML}{4077ab}
\definecolor{red}{HTML}{CC8E7F}
\definecolor{magenta}{HTML}{A748C3}

\newcommand{\eat}[1]{\ignorespaces}
\newcommand{\xxcomment}[4]{\textcolor{#1}{[$^{\textsc{#2}}_{\textsc{#3}}$ #4]}}
\newcommand{\ben}[1]{\xxcomment{blue}{B}{A}{#1}}

\newcommand{\D}{\mathcal{D}}

\renewcommand{\S}{\mathcal{S}}
\newcommand{\Q}{\mathcal{Q}}
\newcommand{\bert}{{\sc BERT}\xspace}

\newcommand{\na}{}
\newcommand{\ours}{Ours\xspace}
\newcommand{\tfive}{{\sc t5}\xspace}
\newcommand{\tfivebase}{{\sc t5-base}\xspace}

\newcommand{\snips}{{SNIPS}\xspace}
\newcommand{\conll}{{CoNLL}\xspace}
\newcommand{\ontonotes}{{Ontonotes}\xspace}
\newcommand{\atis}{{ATIS}\xspace}
\newcommand{\ts}[1]{``#1''\xspace}
\newcommand{\bio}{BIO\xspace}
\newcommand{\bl}[1]{\textbf{#1}}
\newcommand{\Y}{\mathcal{Y}}

\usepackage{tikz}

\title{Augmented Natural Language for Generative Sequence Labeling}

\author{Ben Athiwaratkun \\
  AWS AI \\
  \texttt{benathi@amazon.com} \\ \And
  Cicero Nogueira dos Santos \\
  AWS AI \\
  \texttt{cicnog@amazon.com} \\ \AND
  Jason Krone \\
  AWS AI \\
  \texttt{kronej@amazon.com} \\\And
  Bing Xiang \\
  AWS AI \\
  \texttt{bxiang@amazon.com} \\
  }

\date{}

\begin{document}

\maketitle
%\notice
\begin{abstract}
We propose a generative framework for joint sequence labeling and sentence-level classification. 
Our model performs multiple sequence labeling tasks at once using a single, shared natural language output space. 
Unlike prior discriminative methods, our model naturally incorporates label semantics and shares knowledge across tasks.
Our framework is general purpose, performing  well on few-shot, low-resource, and high-resource tasks.
We demonstrate these advantages on popular named entity recognition, slot labeling, and intent classification benchmarks. We set a new state-of-the-art for few-shot slot labeling, improving substantially upon the previous 5-shot ($75.0\% \rightarrow 90.9\%$) and 1-shot ($70.4\% \rightarrow 81.0\%$) state-of-the-art results. Furthermore, our model generates large improvements ($46.27\% \rightarrow 63.83\%$) in low-resource slot labeling over a \bert baseline by incorporating label semantics. 
We also maintain competitive results on high-resource tasks, % as well, 
performing within two points of the state-of-the-art on all tasks and setting a new state-of-the-art on the \snips dataset.
% BenA: people might be curious why the low resource numbers are lower -- because of higher amount of classes
% also it's not a standard evaluation -- perhaps we cut the numbers for low-resource
\end{abstract}

\section{Introduction} \label{sec:intro}
Transfer learning has been the pinnacle of recent successes in natural language processing. Large pre-trained language models are powerful backbones that can be fine-tuned for different tasks to achieve state-of-the-art performance in wide-raging applications \citep{elmo, bert, gpt2, bart, xlnet, roberta}. % a lot more

While these models can be adapted to perform many tasks, each task is often associated to its own output space, which limits the ability to perform multiple tasks at the same time. 
For instance, a sentiment analysis model is normally a binary classifier that decides between class labels \emph{``positive''} and \emph{``negative''}, while a multi-class entailment system classifies each input as \emph{``entail''}, \emph{``contradict''}, or \emph{``neither''}.
This approach presents difficulty in knowledge sharing among tasks. 
That is, to train the model for a new task, the top-layer classifier is replaced with a new one that corresponds to novel classes. 
The class types are specified implicitly through different indices in the new classifier, which contain no prior information about the label meanings.
This discriminative approach does not incorporate label name semantics and often requires a non-trivial amount of examples to train \citep{mixout}.
While this transfer learning approach has been immensely successful, a more efficient approach should incorporate prior knowledge when possible.

Conditional generative modeling is a natural way to incorporate prior information and encode the output of multiple tasks in a shared predictive space. 
Recent work by \citet{t5} built a model called \tfive to perform multiple tasks at once using natural language as its output. 
The model differentiates tasks by using prefixes in its input such as \emph{``classify sentiment:''}, \emph{``summarize:''}, or \emph{``translate from English to German:''} and classify each input by generating natural words such as \emph{``positive''} for sentiment classification or \emph{``This article describes ...''} for summarization.

\begin{figure}[]
\begin{center}
\centerline{\includegraphics[width=\columnwidth,trim={5 265 5 180},clip]{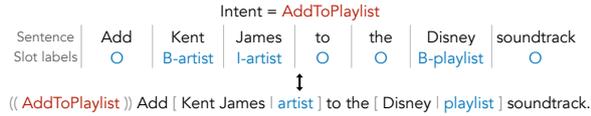}}
\caption{
The conversion between the canonical  \bio tagging format and our augmented natural language format.
}
\label{fig:example_conversion}
\end{center}
%\vskip -0.4in
\end{figure}

However, the appropriate output format for important \emph{sequence labeling} applications in NLP, such as named entity recognition (NER) and slot labeling (SL) is not immediately clear. 
In this work, we propose an \emph{augmented natural language} format for sequence labeling tasks. Our format locally tags words within the  sentence (Figure \ref{fig:example_conversion}) and is easily extensible to sentence-level classification tasks, such as intent classification (IC).

Our highlighted contributions and main findings are as follows:
\begin{enumerate}[1)]
\itemsep-0.2em 
    \item We propose an effective new output format to perform joint sequence labeling and sentence classification through a generation framework.
    \item We demonstrate the ability to perform multiple tasks such as named entity recognition, slot labeling and intent classification within a single model.
    \item Our approach is highly effective in low-resource settings. Even without incorporating label type semantics as priors, the generative framework learns more efficiently than a token-level classification baseline. 
    % due to the robustness of sequence to sequence model to overfitting, as well as stronger attention. \ben{verify claim please!}
    The model improves further given natural word labels, indicating the benefits of rich semantic information.
    \item We show that supervised training on related sequence labeling tasks acts as an effective meta-learner that prepares the model to generate the appropriate output format. 
    Learning each new task becomes much easier and results in significant performance gains.
   \item We set a new state-of-the-art for few-shot slot labeling, outperforming the prior state-of-the-art by a large margin. 
\item We plan to open source our implementation and will update the paper with a link to our repository.
%   \item We open source our implementation\footnote{Implementation to be released upon publication.} to support future work on sequence labeling.
%Amazon internal version can be accessed at {\tiny \url{https://code.amazon.com/packages/Meta-sequence-label/trees/mainline}}

   \end{enumerate}
%\vskip -0.09in

\begin{figure*}[ht]
\begin{center}
\centerline{\includegraphics[width=\linewidth,trim={20 260 8 165},clip]{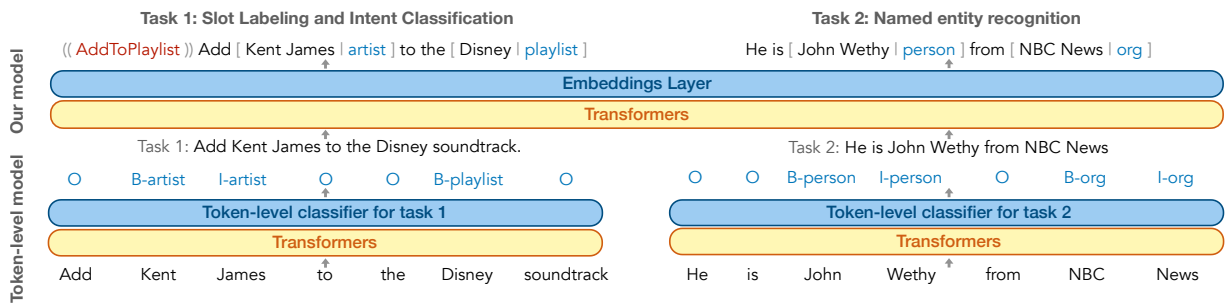}}
\caption{
Comparison between our generative-style sequence labeling model (top) and the conventional token-level classification model (bottom). 
}
\label{fig:model_descriptions}
\end{center}
%\vskip -0.32in
\end{figure*}

\section{Model} \label{sec:model}

\subsection*{Sequence Labeling as Generation}
Most work on sequence labeling uses token-level classification frameworks. That is, given a list of tokens $\ell = \{ \ell_i \}_{i=1}^n$, we perform a prediction on every token $\ell_i$ to obtain $y' = \{y'_i\}_{i=1}^n =  \{ f(\ell_i;\ell)\}_{i=1}^n$ where $f(\cdot)$ is a token-level prediction function. The prediction is accurate if it matches the original sequence label $y = \{ y_i \}_{i=1}^n$.

In contrast to this convention, we frame sequence labeling as a conditional sequence generation problem where given the token list $\ell$, we generate an output list $o = g(\ell)$ where $g$ is a sequence-to-sequence model. 
A ``naive'' formulation for this task would be to directly generate $o=y$ given $\ell$.
However, this approach is prone to errors such as word misalignment and length mismatch (see supplementary materials Section \ref{supp:generate_list} for discussion).

We propose a new formulation for this generation task such that, given the input sequence $\ell$, our method generates output $o$ in \textbf{augmented natural language}. The augmented output $o$ repeats the original input sequence $\ell$ with additional markers that indicate the token-spans and their associated labels. 
More specifically, we use the format { [ $\ell_j, \hdots, \ell_{j+t}$ \textvertline \ { \color{black} $L$} ]} to indicate that the token sequence $\ell_j, \hdots, \ell_{j+t}$ is labeled as $L$. 

Fig. \ref{fig:example_conversion} depicts the proposed format and its equivalent canonical \bio format for the same input sentence.
The conversion between the \bio format and our augmented natural language format is invertible without any information loss.
This is crucial so that the generated output from model prediction can be converted back for comparison without uncertainty.

There are other formats which can encapsulate all the tagging information but are not invertible. For instance, outputting only the token spans of interest with tagging patterns { [ $\ell_j, \hdots, \ell_{j+t}$ \textvertline \ { \color{black} $L$} ]} without repeating the entire sentence results in the invertibility breaking down when there are duplicate token spans with different labels. We discuss this further in the appendix Section \ref{supp:example_f3}.

\subsection*{Joint Sequence Classification and Labeling}
Our sequence to sequence approach also supports joint sentence classification and sequence labeling by incorporating the sentence-level label in the augmented natural language format. 
%It is essentially a matter of incorporating a sentence-level label to the generated output.
In practice, we use the pattern {(( sentence-level label ))} in the beginning of the generated sentence, as shown in Fig. \ref{fig:example_conversion}. 
The use of double parentheses is to prevent confusion with a single parenthesis that can occur in the original word sequence $\ell$.

\subsubsection*{Training and Evaluation}
We train our model by adapting the pre-trained \tfive with the sequence to sequence framework. 
Additionally, we prefix the input with task descriptors in order to simultaneously perform multiple classification and labeling tasks, similar to the approach by \citet{t5}. 
This results in a seamless multi-task framework, as illustrated in the top part of Fig. \ref{fig:model_descriptions}.
To evaluate, we convert the generated output back to the canonical BIO format and calculate the F1 score for sequence labeling or accuracy for sentence classification.

\subsection*{Natural Labels}
Labels are associated to real-world concepts that can be described through natural words.
These words have rich information, but are often ignored in traditional discriminative approaches. 
In contrast, our model naturally incorporate label semantics directly through the generation-as-classification approach.

We perform label mapping in order to match the labels to its natural descriptions and use the natural labels in the augmented natural language output. Our motivation is as follows: 
(1) Pre-trained conditional generation models which we adapt on have richer semantics embedded in natural words, rather than dataset-specific label names. For instance, ``country city state'' contains more semantic information compared to ``GPE'', which is an original label in named entity recognition tasks.
Using natural labels should allow the model to learn the association between word tokens and labels more efficiently, without requiring many examples. 
(2) Label knowledge can be shared among different tasks. For instance, after learning how to label names as \ts{person}, given a new task in another domain which requires labeling \ts{artist}, the model can more easily associate names with \ts{artist} due to the proximity of \ts{person} and \ts{artist} in embeddings. This is not the case if the concept of \ts{person} was learned with other uninformative words.

% \rotatebox{90}{B} -- 
\begin{table*}[h]
\begin{center}
\begin{tabular}{p{0.2cm} p{6.1cm}  c c c c c c c c c c }
 & \multirow{2}{*}{\textbf{Task \& Dataset}}                                        & \multicolumn{2}{c}{\textbf{Intent Clas.}} & \multicolumn{4}{c}{\textbf{Slot Labeling}}                             \\
\cmidrule(lr){3-4} \cmidrule(lr){5-8}
  &                                      & \textbf{SNIPS}           & \textbf{ATIS}           & \textbf{SNIPS} & \textbf{ATIS}  & \textbf{\conll} & \textbf{Onto} \\ %\hline
%No. of intent types                          & 7                        & 21                      & -              & -              & -              & -                  \\
%No. of slot types                            & -                        & -                       & 39             & 83             & 4              & 18                 \\
% & No. of label types 			  & 7                        & 21  		 & 39             & 83             & 4              & 18                 \\
% & No. of training sentences                    & 13084                    & 4478                    & 13084          & 4478           & 14041          & 59924              \\ \cmidrule(lr){2-8}
\multirow{3}{*}{ \rotatebox{90}{SL/IC}  } 
 & Bi-Model \citep{bimodel}                     & \na                       & \textbf{98.99}          & \na             & \textbf{96.89}          & \na             & \na                 \\
% & Stack-Prop \citep{stackprop}          & 98.00                    & 96.90                   & 94.2           & 95.90          &  \na             &    \na               \\
% & Stack-Prop w Oracle \citep{stackprop} & \na                       & \na                      & 96.10          & 96.00          &   \na            &     \na              \\
 & Joint \bert \citep{jointbert_icsl}            & 98.60                    & 97.50                   & \textbf{97.00}          & 96.10          & \na             & \na                 \\
 & ELMO+BiLSTM \citep{bilstm_crf_elmo}  & \textbf{99.29}           & 97.42                   & 93.90          & 95.62          &    \na           &     \na              \\
% & Capsule Networks \citep{capsule_icsl}        & 97.70                    & 95.00                   & 91.80          & 95.20          &    \na           &   \na                \\ 
 %\cmidrule(lr){2-8} 
 [0.12cm]
 \multirow{4}{*}{ \rotatebox{90}{NER}  } 
 & Cloze-CNN \citep{cloze_pretrain}  &  \na                      & \na                      & \na             & \na             & \textbf{93.50}          & \na                 \\
 & BERT-MRC \citep{bert_mrc}                    & \na                       & \na                      & \na             & \na             & 93.04          & 91.11              \\
 & BERT-MRC + DSC \citep{dice_loss}             & \na                       & \na                      & \na             & \na             & 93.33          & \textbf{92.07}              \\
% & BERT Large \citep{bert}                      & \na                       & \na                      & \na             & \na             & 92.80          & \na                 \\
 & BERT Base \citep{bert}                       & \na                       & \na                      & \na             & \na             & 92.40          & 88.95                  \\ 
% & Hierarchical BERT \citep{hier_bert}          & \na                       & \na                      & \na             & \na             & 93.37          & 90.30              \\
% & Flair Embeddings \citep{flair_embeddings}    & \na                       & \na                      & \na             & \na             & 93.09          & 89.30              \\ 
\cmidrule(lr){2-8} 
% & SOTA Summary                                 & 99.29                    & 99.10                   &   97.00       & 96.89           & 93.50          & 92.07              \\ 
% & Ours: \snips                                  & 99.00                    & \na                      & \underline{\textbf{97.43}} & \na             & \na             & \na                 \\
% & Ours:\atis                                   & \na                       & 96.86                   & \na             & 96.13          & \na             & \na                 \\
% & Ours: \conll                                  & \na                       & \na                      & \na             & \na             & 90.70          & \na                 \\
% & Ours: \ontonotes                              & \na                       & \na                      & \na             & \na             & \na             & \textbf{90.24}     \\
 & Ours: Individual   & 99.00 & 96.86 & \underline{\textbf{97.43}} & 96.13 & 90.70 & \textbf{90.24} \\ 
 & Ours: \snips{}+\atis                          & \underline{\textbf{99.29}}           & \textbf{97.20}          & 97.21          & 95.83          &      \na         &     \na              \\
 & Ours: \conll{}+\ontonotes                      & \na                       & \na                      & \na             & \na     & \textbf{91.48} & 89.52              \\
%Ours: \snips +\atis + \conll                   & 99.00                    & 97.09                   & 96.35          & 95.37          & 90.89          &                    \\
 & Ours: \snips{}+\atis{}+\conll{}+\ontonotes       & 99.14                    & 97.08                   & 96.82          & \textbf{96.65} & \textbf{91.48} & 89.67 \\
%Ours w t5-large: \snips +\atis + \conll + \ontonotes       &  99.29                   &    97.09                &    97.47       & 96.19 & \textbf{91.90} & 89.75 \\
\end{tabular}
\caption{Results of our models trained on combinations of datasets. Results for \textbf{Ours: individual} are from models trained on a single respective dataset. 
We underline scores of our models that exceed previous state-of-the-art results in each domain. Scores in boldface are the best overall scores among our models, or among the baselines. We use the boldface and underline notation for the rest of the paper. 
} \label{tab:main}
\end{center} 
%\vskip -0.15in
\end{table*}

\section{Related Work} \label{sec:related}

Sequence to sequence learning has various applications including machine translation \citep{seq2seq, seq2seq_att}, language modeling \citep{gpt1, t5}, abstractive summarization \citep{rush-etal-2015-neural}, generative question answering \citep{unilm}, to name a few. 
However, the sequence-to-sequence framework is often not a method of choice when it comes to sequence labeling. 
Most models for sequence labeling use the token-level classification framework, where the model predicts a label for each element in the input sequence \citep{cloze_pretrain, dice_loss, jointbert_icsl}.
While select prior work adopts the sequence-to-sequence method for sequence labeling \citep{chen-moschitti-2018-learning}, this approach is not widely in use due to the difficulty of fixing the output length, output space, and alignment with the original sequence. 

Multi-task and multi-domain learning often benefit sequence labeling performance  \citep{changpinyo-etal-2018-multi}. The archetypal multi-task setup jointly trains on a target dataset and one or more auxiliary datasets. 
In the cross lingual setting, these auxiliary datasets typically represent high-resource languages \citep{schuster2018cross, cotterell-duh-2017-low}. While in a monolingual scenario, the auxiliary datasets commonly represent similar, high-resource tasks. Examples of similar multi-task pairs include NER and slot labeling \citep{louvan-magnini-2019-leveraging} as well as dialogue state tracking and language understanding \citep{rastogi-etal-2018-multi}.

A recent series of works frame natural language processing tasks, such as translation, question answering, and sentence classification, as conditional sequence generation problems \citep{t5, gpt2, gpt3}. By unifying the model output space across tasks to consist of natural language symbols, these approaches reduce the gap between language model pre-training tasks and downstream tasks. Moreover, this framework allows acquisition of new tasks without any architectural change. The GPT-3 model \citep{gpt3} demonstrates the promise of this framework for few-shot learning. Among other successes, GPT-3 outperforms \bert-Large on the SuperGLUE benchmark using only 32 examples per task.   To the best of our knowledge, we are the first to apply this multi-task conditional sequence generation framework to sequence labeling.

The conditional sequence generation framework makes it easy to incorporate label semantics, in the form of label names such as \textit{departure city}, example values like \textit{San Francisco}, and descriptions like \textit{``the city from which the user would like to depart on the airline''}. Label semantics provide contextual signals that can improve model performance in multi-task and low-resource scenarios. Multiple works show that conditioning input representations on slot description embeddings improves multi-domain slot labeling performance \citep{bapna2017towards, lee2019zero}.  
Embedding example slot values in addition to slot descriptions yields further improvements in zero-shot slot labeling \citep{shah-etal-2019-robust}.
In contrast to our work, these approaches train slot description and slot value embedding matrices, whereas our framework can incorporate these signals as natural language input without changing the network architecture.

\section{Experimental Setup and Results} \label{sec:experiments}

\subsection{Data} \label{sec:data}
\paragraph{Datasets} We use popular benchmark data \snips \citep{snips} and \atis \citep{atis} for slot labeling and intent classification. \snips is an SLU benchmark with 7 intents and 39 distinct types of slots, while \atis is a benchmark for the air travel domain (see appendix \ref{sec:supp_data} for details). 
We also evaluate our approach on two named entity recognition datasets, \ontonotes \citep{ontonotes} and \conll-2003 \citep{conll}.

\paragraph{Construction of Natural Labels}
We preprocess the original labels to natural words as follows. For  \ontonotes and \conll datasets, we transform the original labels via mappings
detailed in Table \ref{tab:ontonotes_labelmap} and \ref{tab:conll_labelmap} in the appendix.
For instance, we map \ts{PER} to \ts{person} and \ts{GPE} to \ts{country city state}.  For \snips and \atis, we use the following rules to convert intent and slot labels: (1) we split words based on ``.'', ``\textunderscore'', ``/", and capitalized letters. For instance, we convert \ts{object{\textunderscore}type} to \ts{object type} and \ts{AddToPlaylist} to \ts{add to playlist}. These rules result in better tokenization and enrich the label  semantics. 
We refer to these as the \textbf{natural label} setting and use is as our default.

\subsection{Multi-Task Sequence Classification and Slot Labeling} \label{sec:main_exps}
We first demonstrate that our model can perform multiple tasks in our generative framework and achieve highly competitive or state-of-the-art performance.
We consider 4 sequence labeling tasks and 2 classification tasks: 
NER on  \ontonotes and \conll datasets; and slot labeling (SL) and intent classification (IC) on  \snips and \atis dialog datasets. 
For comparison, we provide baseline results from the following models:

\textbf{SL and IC:}
\textbf{Bi-Model} \citep{bimodel} uses two correlated bidirectional LSTMs to perform both IC and SL.
\textbf{Joint BERT} \citep{jointbert_icsl} performs joint IC and SL with a sequential classifier on top of BERT, where the classification for the start-of-sentence token corresponds to intent class. 
\textbf{ELMO+Bi-LSTM} \citep{bilstm_crf_elmo}  uses a Bi-LSTM with CRF as a classifier on top of pre-trained ELMO \citep{elmo}.

\textbf{NER:}
\textbf{Cloze-CNN} \citep{cloze_pretrain} fine-tunes a Bi-LSTM with CRF model \citet{elmo} on a pre-trained model with a cloze-style word reconstruction task.
\textbf{BERT MRC} \citep{bert_mrc} performs sequence labeling in a question answering model to predict the slot label span. 
\textbf{BERT MRC + Dice Loss} \citep{dice_loss} improves upon \bert MRC with a dice loss shown to be suitable for data with imbalanced labels.
\textbf{BERT} \citep{bert} refers to a token-level classification with \bert pre-trained model. Note that the results for \bert with \ontonotes are from our own implementation. 
%\item \textbf{Hierarchical BERT} \citet{hier_bert} uses both sentence-level and document-level representation.
%\item \textbf{Flair Embeddings} \citet{flair_embeddings} proposes  contextual string embeddings based on character-level models that are suitable for sequential labeling.

\begin{figure*}[]
\centering
  \begin{subfigure}[t]{0.32\textwidth}
    \includegraphics[width=1.\textwidth]{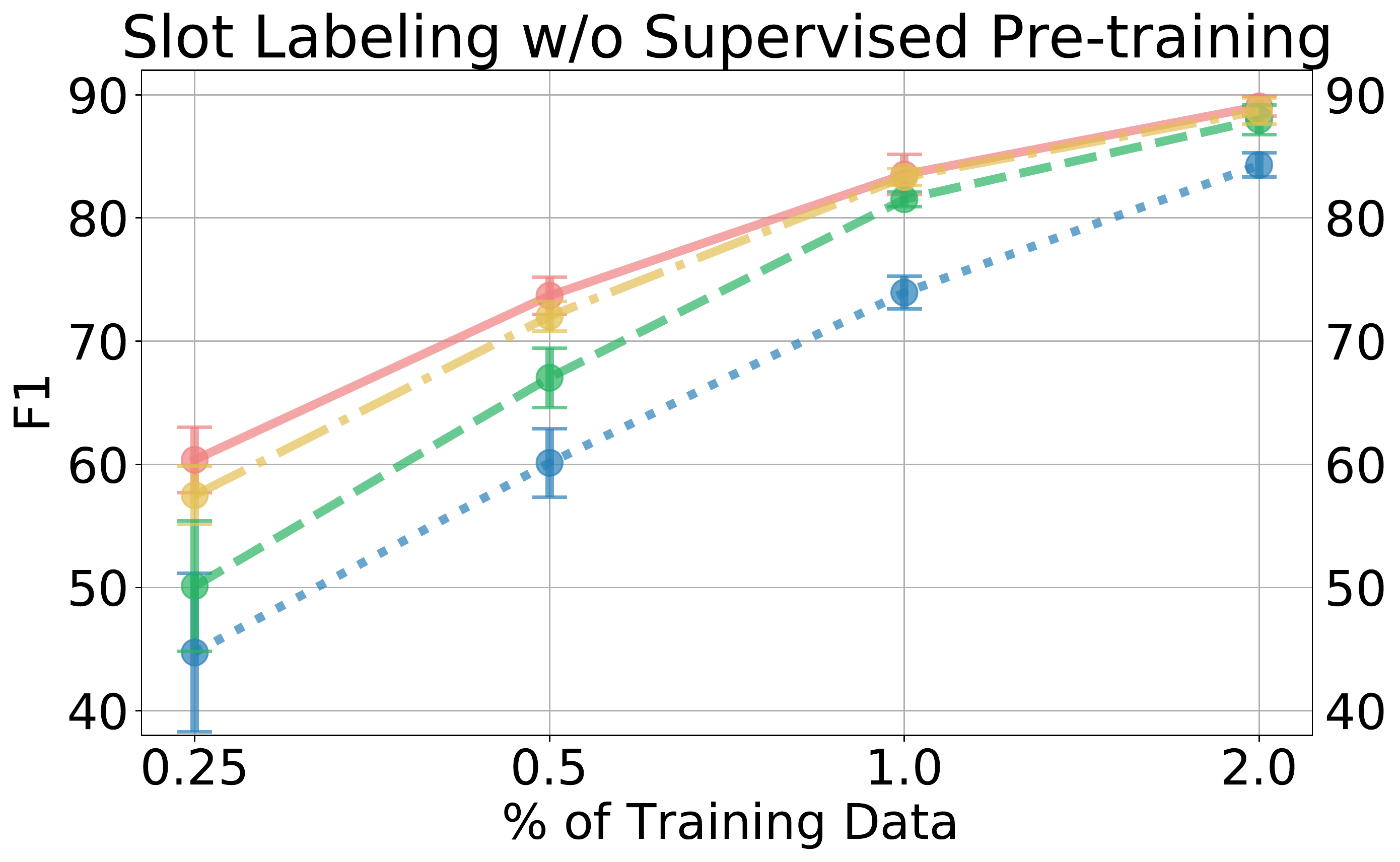}
    \caption{}\label{fig:limited_unsup}
  \end{subfigure}
  \begin{subfigure}[t]{0.32\textwidth}
    \includegraphics[width=1.\textwidth]{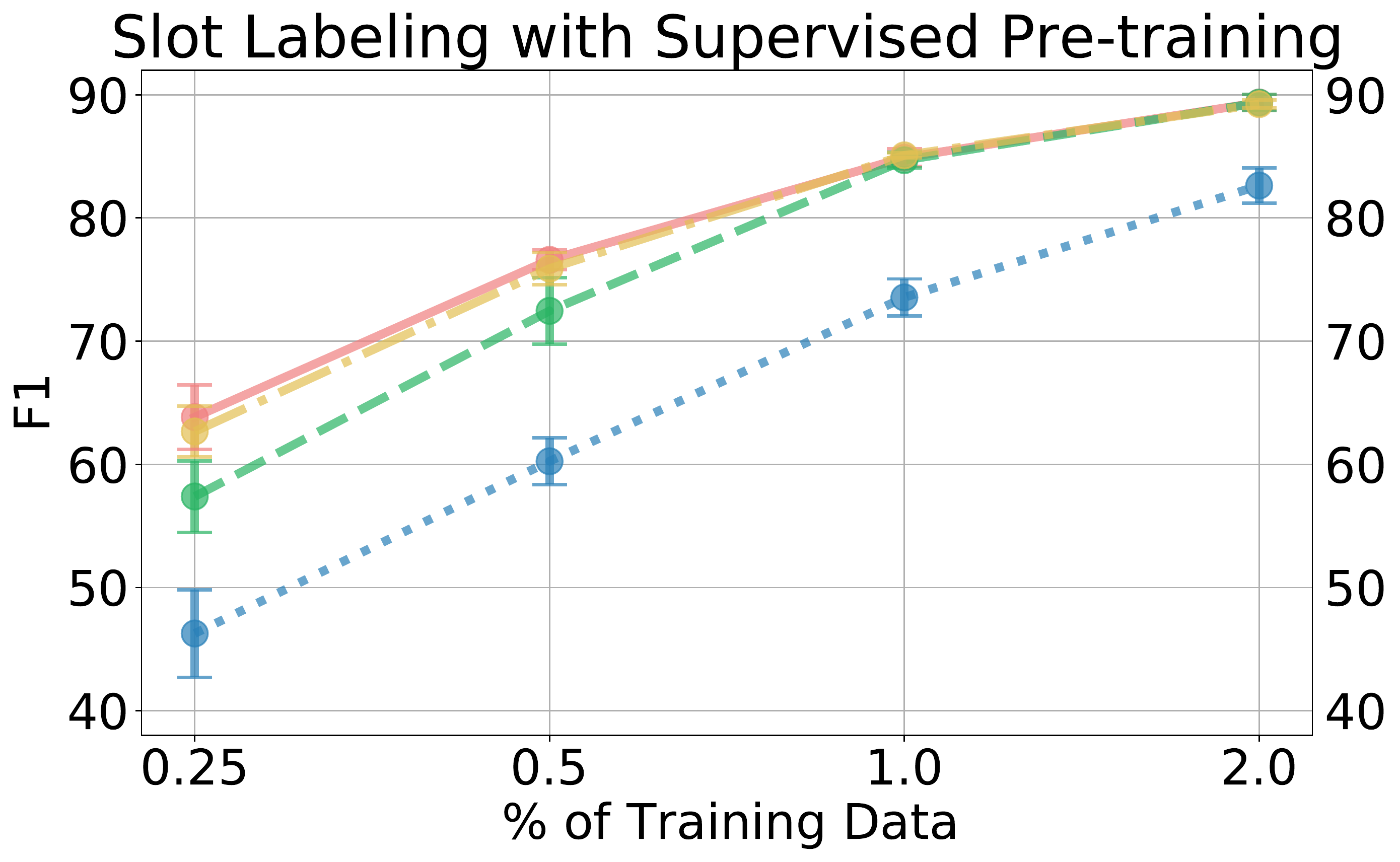}
    \caption{}\label{fig:limited_sup}
  \end{subfigure}
  \begin{subfigure}[t]{0.32\textwidth}
    \includegraphics[width=1.\textwidth]{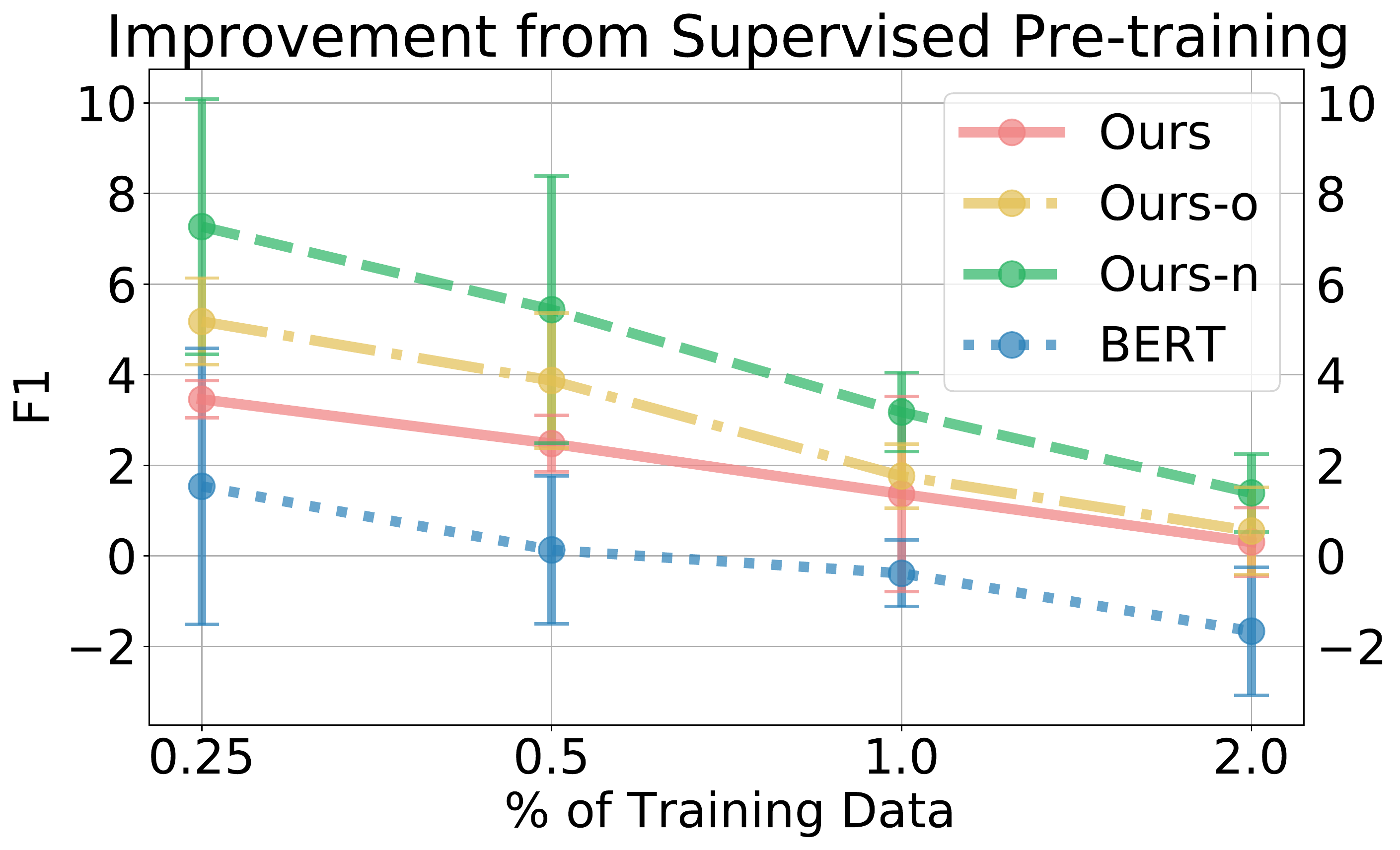}
     \caption{}\label{fig:limited_diff}
  \end{subfigure}
  \vspace{-0.5\baselineskip}
  \caption{
Model performance on limited amount of training data. The error bars indicate the standard deviation over $4$ random trials. 
\textbf{\ours-o} is our model with its original labels. \textbf{\ours-n} is our model with numeric labels. 
  }
  \label{fig:limited}
  %\vskip -0.15in
\end{figure*}

In Table \ref{tab:main}, we report a summary of the results for our method and the baselines.
Our proposed model achieves highly competitive results for \atis, \ontonotes, and \conll datasets, as well as state-of-the-art slot labeling and intent classification performance on the \snips dataset.
Unlike all the baseline models, which can perform a single task on a specific dataset, our model can perform all the tasks considered at once (last row of Table \ref{tab:main}).
For the multi-task models, our results show that different sequence labeling task can help mutually benefit each other, where \atis slot labeling result improves from 96.13 to 96.65 and \conll improves from 90.70 to 91.48.
While there are other approaches that perform better than our models in some tasks, we highlight the simplicity of our generation framework which performs multiple tasks seamlessly.
This ability helps the models transfer knowledge among tasks with limited data, which are demonstrated through the rest of the paper.

\subsection{Limited Resource Scenarios and Importance of Label Semantics} \label{sec:limited_resource} \label{sec:ablation}
In this section, we show that our model can use the semantics of labels to learn efficiently, which is crucial for scenarios with limited labeled data. 
To demonstrate this effect, we use our model with the following variants of labels which differ semantic quality: (1) \textbf{natural label}, (2) \textbf{original label} and (3) \textbf{numeric label}.

The natural label version is our default setting where we use labels expressed in natural words.
The original label case uses labels provided by the datasets, and the numeric label case uses numbers $0,1,2, ...$ as label types. 
In the numeric version, the model does not have pre-trained semantics of the label types and has to learn the associations between the labels and the relevant words from scratch. 
We also compare with the \textbf{\bert token-level classification} model.  Similar to the numeric label case, the label types for \bert do not initially have associated semantics and are implicit through indices in the classifier weights. 
We use the \snips dataset to conduct our experiments due to its balanced domains (see  Table \ref{tab:snips_stats} in Appendix). 
We experiment with very limited resource scenarios where we use as low as $0.25\%$ of training data, corresponding to roughly one training sentence per label type on average.

Figure \ref{fig:limited_unsup} shows the sequence labeling performance for varying amount of training data (see Table \ref{tab:snips_lowresource} in the appendix for numeric results). 
We observe that label semantics play a crucial role in the model's ability to learn effectively for limited resource scenarios. 
Our model with natural labels outperforms all other models, achieving an F1 score of $60.4 \pm 2.7\%$ with $0.25\%$ training data, and giving a slight boost over using original labels ($57.5\pm2.4\%$). 
We believe that the improvement can be more dramatic in other datasets where the original labels have no meanings (such as in the numeric case), are heavily abbreviated, or contain rare words.
%By using the numeric labels in place of natural labels, we prevent the models from using prior label knowledge to learn new tasks.
With the numeric model, the performance suffers significantly in low-resource settings, achieving only $50.1\pm5.3\%$, or $10.3\%$ lower than the natural label model, with $0.25\%$ data.
This result further supports the importance of label semantics in our generation approach.
Interestingly, we also observe that the numeric model still outperforms \bert token-level classification ($44.7\pm6.4\%$), where neither model contains  prior label semantics.
This result indicates that even in the absence of label meanings, the generation approach seems more suitable than the token-level framework.

\begin{comment}
\begin{table}[]
\begin{center}
\begin{tabular}{r c c c c c}
        & \textbf{Ours} & \textbf{Ours$^\dagger$} & \textbf{Ours$^\ddagger$} & \textbf{BERT} \\ \cmidrule(lr){2-5}
w/o sup & 73.69                   & 72.02                    & 67.03                    & 60.12                    \\
  $\pm$      & 1.51                    & 1.22                     & 2.41                     & 2.77                     \\
w sup   & 76.60                   & 73.39                    & 72.46                    & 60.25                    \\
  $\pm$      & 0.79                    & 2.09                     & 2.70                     & 1.89                    
\end{tabular}
\caption{Model performance with $0.5\%$ training data with and without supervised pre-training. We use $4$ random trials and report the mean of F1 and its standard deviation. \ben{see if this table is necessary}
} \label{tab:limited_mini}
\end{center}
\end{table}
\end{comment}

\subsection{Teaching Model to Generate via Supervised Transfer Learning} \label{sec:supervised_transfer_learning}

While we train our model in limited data scenarios, we are asking the model to generate a new format of output given small amount of  data. 
This is challenging since a sequence generation framework typically requires large amount of training \citep{seq2seq}. 
Despite this challenge, our model is able to outperform the classical token-level framework with ease.
This section explores a clear untapped potential -- by teaching our model how to generate the augmented natural language format before adapting to new tasks, we show that the performance on limited data significantly improves.
This result  contrasts with the \bert token-level model where supervised transfer learning hurts overall performance compared to \bert's initial pre-training due to possible overfitting.

To conduct this experiment, we train our model with the \ontonotes NER task in order to teach it the expected output format. 
Then, we adapt it on another task (SNIPS) with limited data, as in Section \ref{sec:limited_resource}.
We compare the results with the token-level \bert model, which also uses the \bert model trained on \ontonotes for supervised pre-training. 
We demonstrate the results in Figure \ref{fig:limited_sup} as well as highlight the improvement due to supervised pre-training in Figure \ref{fig:limited_diff}. We also provide full numeric results in the appendix Table \ref{tab:snips_lowresource_suppretrain_ner} for reference.

Our model demonstrates consistent improvement, achieving an F1 score of $63.8 \pm 2.6\%$ using $0.25\%$ of the training dataset, compared to $60.4 \pm 0.27\%$ without supervised transfer learning. 
The improvement trend also continues for other data settings, as shown in Figure \ref{fig:limited_diff}. 
The benefits from transfer learning is particularly strong for the numeric label model, achieving  $57.4 \pm 2.9\%$ compared to $50.1 \pm 5.3\%$ for $0.25\%$ data.
This results suggests that the initial knowledge from supervised pre-training helps the model associate its labels (without prior semantics) to the associated words more easily.

The supervised transfer learning can also be seen as a \emph{meta-learner}, which teaches the model how to perform sequence labeling in the generative style.
In fact, when we investigate the model output \emph{without} adapting to the \snips dataset, in addition to the output having the correct format, it already contains relevant tagging information for new tasks.

For instance, a phrase ``Onto jerry’s Classical Moments in Movies'' from the \snips dataset results in the model output ``Onto jerry’s {\color{gray} [} Classical Moments in Movies {\color{gray} \textvertline} {\color{blue} \it work of art} {\color{gray} ]}''.
This prediction closely matches the true label ``Onto {\color{gray} [} jerry’s {\color{gray} \textvertline} {\color{blue} \it playlist owner} {\color{gray} ]} {\color{gray} [} Classical Moments in Movies {\color{gray} \textvertline}  {\color{blue} \it playlist} {\color{gray} ]}'' where the true class of ``Classical Moments in Movies'' is {\color{blue} \it playlist}.
Intuitively, the classification as {\color{blue} \it work of art} is in agreement with the true label {\color{blue} \it playlist}, but simply needs to be refined to match the allowed labels for the new task.

In contrast to our framework where the supervised transfer learning helps teach the model an output style, the transfer learning for the token-level classification simply adapts its weights and retains the same token-level structure (albeit with a new classifier). 
We observe no significant improvement from supervised pre-training for the \bert token-level model, which obtains an F1 score of $46.3 \pm 3.6\%$ compared to $44.7 \pm 6.4\%$ without supervised pre-training (with $0.25\%$ SNIPS data).
The improvements are also close to zero or negative for higher data settings (Figure \ref{fig:limited_diff}), suggesting that the pre-training of the token-level classification might overfit to the supervised data, and results in lower generalization on other downstream tasks.
Overall, the final result on the \bert model lags far behind our framework, performing $17.5\%$ lower than our model's score for  $0.25\%$ training data.

In addition, our model with numeric labels performs much better than the \bert token-level model and further highlights the suitability of our generative output format for sequence labeling, regardless to the label semantics. 
Possible explanations are that the sequence to sequence label is less prone to overfitting compared to the classification framework. 
It could also be the case that locally tagging words with labels in the word sequence helps improve attention within the transformers model, and improve robustness to limited data.

\begin{table*}[]
\begin{center}
\begin{tabular}{p{0.2cm} lcccccccccccc}
&        & \textbf{We} & \textbf{Mu} & \textbf{Pl} & \textbf{Bo} & \textbf{Se} & \textbf{Re} & \textbf{Cr} & \textbf{Ave.} \\ \cmidrule(lr){2-10}
 \multirow{8}{*}{\rotatebox{90}{1-shot} }
%                              & SimBERT            & 36.10                & 37.08                & 35.11                & 68.09                & 41.61                & 42.82                & 23.91                & 40.67                     \\
                              & TransferBERT       & 55.82                & 38.01                & 45.65                & 31.63                & 21.96                & 41.79                & 38.53                & 39.06                     \\
                              & MN + \bert       & 21.74                & 10.68                & 39.71                & 58.15                & 24.21                & 32.88                & \bl{69.66}       & 36.72                     \\
                              & WPZ + BERT         & 46.72                & 40.07                & 50.78                & 68.73                & 60.81                & 55.58                & 67.67                & 55.77                     \\
                              & L-TapNet+CDT       & \bl{71.53}       & \bl{60.56}       & \bl{66.27}       & \bl{84.54}       & \bl{76.27}       & \bl{70.79}       & 62.89                & \bl{70.41}            \\ 
\cmidrule(lr){2-10}
                              & Ours + \snips       & {\ul \textbf{82.62}}    & {\ul \textbf{77.46}}                & {\ul \textbf{71.33}}                & {\ul \textbf{85.49}}                & {\ul \textbf{83.22}}                & {\ul \textbf{84.23}}                & {\ul \textbf{82.92}}                & {\ul \textbf{81.04}}                     \\
                              & Ours + Onto        & 56.39                & {\ul 67.10}                & 53.49                & 71.94                & 66.21                & 69.04                & 28.80                & 59.00                     \\
                              & Ours + No Meta          & 46.42                & 59.02                & 47.47                & 63.79                & 49.42                & 64.45                & 17.60                & 49.74                     \\
[0.3cm] 
\multirow{8}{*}{\rotatebox{90}{5-shot} }
%& SimBERT      & 53.46                  & 54.13                  & 42.81                  & 75.54                  & 57.10                  & 55.30                  & 32.38                  & 52.96                    \\
& TransferBERT & 59.41                  & 42.00                  & 46.70                  & 20.74                  & 28.20                  & 67.75                  & 58.61                  & 46.11                    \\
& MN + BERT& 36.67                  & 33.67                  & 52.60                  & 60.09                  & 38.42                  & 33.28                  & 72.10                  & 47.98                    \\
& WPZ + BERT   & 67.82                  & 55.99                  & 46.02                  & 72.17                  & 73.59                  & 60.18                  & 66.89                  & 63.24                    \\
& L-TapNet+CDT & \bl{71.64}                  & \bl{67.16}                  &  \bl{75.88}                  &  \bl{84.38}                  & \bl{82.58}                  & \bl{70.05}                  & \bl{73.41}                  & \bl{75.01}   \\ %     \vspace{0.25cm}             \\ %
\cmidrule(lr){2-10}
& \ours + \snips &{\ul \textbf{91.35}} & {\ul \textbf{86.73}} & {\ul \textbf{87.20}} & {\ul \textbf{95.85}} & {\ul \textbf{92.71}} & {\ul \textbf{91.23}} & {\ul \textbf{91.55}} & \multicolumn{1}{r}{{\ul \textbf{90.95}}} \\
& \ours + Onto  & {\ul 83.15} & {\ul {86.15}} & {\ul {80.36}} & {\ul {90.27}} & {\ul {84.87}} & {\ul {85.89}} & 68.08 & {\ul {82.68}} \\
& \ours + No Meta  & {\ul 73.14}          & {\ul 82.02}          & {\ul 78.82}          & {\ul 84.86}          & {\ul 83.14}          & {\ul 86.63}          & 52.56 & {\ul 77.31}          \\
\end{tabular}
\caption{Our few-shot slot labeling results on $7$ domains of \snips dataset. 
%We underline scores of our models that exceed previous state-of-the-art results. Scores in boldface are the best overall scores. 
\ours + SNIPS perform meta-training on the leave-one-out \snips data, similar to other baselines. 
\ours + Onto is our model trained on \ontonotes. \ours + No Meta involves no meta-training. 
} \label{tab:few_shot}
\end{center} 
%\vskip -0.2in
\end{table*}

\subsection{Few-Shot Sequence Labeling} \label{sec:few_shot}

\subsubsection{Few-Shot Learning}

In few-shot learning, we seek to train models such that given a new task, the models are able to learn efficiently from few labels. 
Different tasks are sampled from various data domains which differ in terms of allowed labels and other nuances such as input styles.

We define a data domain $\D$ as a set of labeled examples $\D = \{ (x_i, y_i) \}_{i=1}^{N_\D}$ which has its set of allowed label types $\Y_\D \ni y_i $. 
Few-shot learning approaches are evaluated over many \emph{episodes} of data, which represent a variety of novel tasks. 
Each episode $(\S,\Q)$ consists of a support set $\S$ containing $K$-shot labeled samples, as well as a query set $\Q$ used for evaluation.
Data from the evaluation episodes are drawn from the \emph{target} domains $\{\D^T_1, \D^T_2, \hdots \}$, which the model has not previously seen.

To learn such models, we typically have access to another set of domains called the \emph{source} domains $\{\D^S_1, \D^S_2, \hdots \}$, which can be used as the training resources. 
In order to train the model to learn multiple tasks well, many few-shot learning approaches use \emph{meta-learning}, or a learning to learn approach, where the model is trained with many episodes drawn from the source domains in order to mimic the evaluation \citep{matching_net, protonet, relation_net, maml}. 
We refer to this as the \emph{episodic training}.

Another approach, called \emph{fine-tuning}, trains the model on a regular training set from the source domains: $ \cup_m \D^S_m$. Given an episode $(\S, \Q)$ at evaluation time, the model fine-tunes it on the support $\S$, typically with a new classifier constructed for the new task, and evaluates on $\Q$.

\subsubsection{Few-Shot Baselines}

%\noindent 
%\textbf{SimBERT} Given a word $x_i$, the model classifies by finding the most similar word $x^\S_j$ in the support set and predicts $y^\S_j$ as the label of $x_i$. The model uses cosine similarity of \bert's embeddings for similarity measure does not perform meta-training.
\textbf{TransferBERT} trains a token-level classification model by fine-tuning. 
%performs token-level classification by pretraining it on source domains, and finetunes the model on the support set during testing. 
\textbf{Matching Net (MN) + BERT} \citet{matching_net} 
Given a word $x_i$, the model classifies by finding the most similar word $x^\S_j$ in the support set and predicts $y^\S_j$ as the label of $x_i$.
The model also adapts the backbone model with episodic training. 
% compares a token embedding in the query set to the token embeddings of the support set, similar to SimBERT. This approach also adapts the backbone \bert model with episodic training.
\textbf{Warm Proto Zero (WPZ) + BERT} \citet{wpz} uses token-level prototypical network \citep{protonet}, which classifies by comparing a word $x_i$ to each class centroid rather than individual sample embeddings. 
%The model trains the backbone with episodic training, similar to MN. 
\textbf{L-TapNet + CDT} \citet{fewshot_sl_hou} uses a CRF framework and  leverages label semantics in representing labels to calculate emission scores and a collapsed dependency transfer method to calculate transition scores. 
We note that all baselines except for { TransferBERT} uses episodic meta-training whereas { TransferBERT} uses fine-tuning. All baseline results are taken from \citet{fewshot_sl_hou}.

Our model performs fine-tuning with the generation framework. The major difference between our model and a token-level classification model such as { TransferBERT} is that we do not require a new classifier for every novel task during the fine-tuning on the support set. The sequence generation approach allows us to use the entire model and adapt it to new tasks, where the initial embeddings contain high quality semantics and help the model transfer knowledge efficiently.

\subsubsection{K-shot Episode Construction} Traditionally, the support set $\S$ is often constructed in $K$-shot formats where we use only $K$ instances of each label type. In sequence labeling problems, this definition is challenging due to the presence of multiple occurrences or multiple label types in a single sentence. We follow \citet{fewshot_sl_hou} by using the following definition of a K-shot setting: All labels within the task appears at least $K$ times in $\S$ and would appear less than $K$ times if any sentence is removed. We sample $100$ episodes from each domain according to this definition.
Note that \citet{fewshot_sl_hou}'s episodes are similar to ours, but preprocess the sentences by lowercasing and removing extra tokens such as commas (see details in Section \ref{sec:fs_full_results}).
Our model is flexible and can handle raw sentences; we therefore use the episodes from the original \snips dataset without any  modifications.

\subsubsection{Data} We perform few-shot experiments on the $7$ domains $\{\D_1, \hdots, \D_7\}$ of the \snips dataset, namely, Weather (We), Music (Mu), Playlist (Pl), Book (Bo), ScreeningEvent (Se), Restaurant (Re), CreativeWork (Cr). 
To evaluate a model on domain $\D_i$, we meta-train the model on $\D'_i = \{ \D_1, \hdots, \D_7 \} - \D_i$. We refer to this as the \emph{leave-one-out} meta-training sets. All other baselines also use this meta-training data setup.

We note that the training set $\D'_i$ has data distributions that closely match $\D_i$ since they are both drawn from the \snips dataset.
We investigate more challenging scenarios where we use an alternative source as a meta-training set, as well as no meta-training. 
In particular, we choose \ontonotes NER task as the alternative source domain.
The benefits of using this setup is such that it establishes a single meta-trained model 
that works across \emph{all} evaluation domains, which we offer as a challenging benchmark for future research.

\subsubsection{Few-Shot Results}
Table \ref{tab:few_shot} demonstrates the results for few-shot experiments.
Our model outperforms previous state-of-the-art on every domain evaluated.
In the 5-shot case, our model achieves an average F1 score of $90.9\%$, exceeding the strongest baseline by $15.9\%$. 
Even without meta-training, the model is able to perform on par with state-of-the-art models, achieving an F1 score of $77.3\%$ versus $75.0\%$ for the baseline. 
Training on an alternative source (NER task) also proves to be an effective meta-learning strategy, performing better than the best baseline by $7.7\%$.
These results indicate that our model is robust in its ability to learn sequence tagging on target domains that differ from sources.
In the 1-shot case, our model achieves an average F1 score of $81.0\%$, outperforming the best baseline significantly ($10.6\%$ improvement).

We note that the average support sizes are around $5$ to $40$ sentences for the 5-shot case, and one to $8$ sentences for the 1-shot case (see Table \ref{tab:five_shot_all} and \ref{tab:one_shot} for details). 
The results are particularly impressive given that we adapt a large transformer model  based on such limited number of samples. 
In comparison to other fine-tuning approaches such as TransferBERT, our model performs substantially better, indicating that our generative framework is a more data-efficient approach for sequence labeling.

\section{Discussion and Future Work} \label{sec:discussion}

Our experiments  consistently show that the generation framework is suitable for sequence labeling and sets a new record for few-shot learning. 
Our model adapts to new tasks efficiently with limited samples, while incorporating the label semantics expressed in natural words. 
This is akin to how humans learn. 
For instance, we do not learn the concept of ``person'' from scratch in a new task, but have prior knowledge that ``person'' likely corresponds to names, and refine this concept through observations.
The natural language output space allows us to retain the knowledge from previous tasks through shared embeddings, unlike the token-level model which needs new classifiers for novel tasks, resulting in a broken chain of knowledge.

Our approach naturally lends itself to life-long learning. 
The unified input-output format allows the model to incorporate new data from any domain.
Moreover, it has the characteristics of a single, life-long learning model that works well on many levels of data, unlike other approaches that only perform well on few-shot or high-resource tasks.  
Our simple yet effective approach is also easily extensible to other applications such as multi-label classification, or structured prediction via nested tagging patterns.

%\clearpage
\bibliographystyle{acl_natbib}
\bibliography{main}

\clearpage
\appendix
\section{Supplementary Materials} \label{sec:supplementary}
% information about the datasets

\subsection{Experiment Setup} \label{sec:exp_setup}  \label{supp:exp_setup}
We describe the experiment setup for reproducibility in this section. We use Huggingface's \tfivebase conditional generation model as well as their trainer ({\tt transformers.Trainer}) with its default hyperparameters to train all our models. The trainer uses AdamW \citep{adam, adamw} and linear learning rate decay. 
%\ben{provide exact details}
%Note that the in case of very limited data, the real batch size per GPU might be lower due to limited training size. 
\begin{itemize}
\itemsep0em
\item We use 8 V100 GPUs for all our experiments.
\item Maximum batch size per GPU = 8.
\item Maximum sequence length = 128 for all tasks except \ontonotes where we use 175.
\item The number of epochs in multi-task experiments is $50$. 
\item The number of epochs for limited-resource experiments is scaled to have the same number of optimization steps as that of when we use the entire training set. For instance, if we use $10\%$ of the entire training set, we use $500$ epochs on that limited set.
%\item For few-shot experiments, we perform full batch optimization.
\item We perform no preprocessing except for replacing the labels with the natural labels described in Section \ref{sec:data}.
\item We use a library {\tt seqleval}\footnote{\UrlFont https://github.com/chakki-works/seqeval.git} for F1 evaluation which supports many tagging format such as \bio, BIOES, etc. 
\item We use {\tt \small bert-base-multilingual-cased} for  our BERT model. 
\end{itemize}

\paragraph{Task Descriptor Prefixes} 
A task descriptor helps encode information about the allowed set of label types. For instance, \conll allows only 4 types of labels whereas the 18 label types in \ontonotes are more fine-grained.
A task descriptor also help encode the nuances among labels; for example, the \conll  dataset has only one tag type \ts{LOC} for location  whereas \ontonotes differentiate locations with \ts{GPE} (countries, cities, states) and a general \ts{LOC} (non-GPE locations, mountain ranges, bodies of water). 
By specifying a task descriptor, we allow the model to learn the implicit constraints in the data and help it be able to distinguish what task it should perform given an input sentence. 
We use the corresponding prefixes ``SNIPS: ", ``ATIS: ", ``Ontonotes: " or ``CONLL: ". 
We ensure that these prefixes can be tokenized given a pretrained tokenizer properly without the model having to use the unknown token.
We demonstrate in \ref{sec:main_exps} that the prefix tags allow us to perform slot labeling (and intent classification) for different datasets using a single model. For a model trained on only a single dataset, such prefix can be omitted and does not affect the performance.

\subsection{A naive approach for sequence labeling as sequence to sequence} \label{supp:generate_list}
We consider training a sequence-to-sequence model where the output is the sequence of \bio tags. Table \ref{tab:seqtoseq_bio} demonstrates an example with model prediction. We find that the model often outputs predictions that are misaligned with the original sentence slots. This is due to the complex relationships with the tokenizer. For instance, the tokenized version of this sentence (using \tfivebase tokenizer) is of length $25$ whereas the original sentence length is $14$. Learning to map the slot labels to the correct tokens can be challenging. 
\begin{verbatim}
_It, _, a, but, s, _San, chi, h, 
_Rural, _Township, _to, _the,
_northeast, _, ,, _the, _Ku,
ant, u, _area, _of, _Tai, pe,
i, _city
\end{verbatim}

\begin{table}[]
\begin{center}
\begin{tabular}{llll}
   & sentence  & label & prediction \\
0  & It        & O     & O          \\
1  & abuts     & O     & O          \\
2  & Sanchih   & B-GPE & O          \\
3  & Rural     & I-GPE & B-GPE      \\
4  & Township  & I-GPE & I-GPE      \\
5  & to        & O     & I-GPE      \\
6  & the       & O     & O          \\
7  & northeast & O     & O          \\
8  & ,         & O     & O          \\
9  & the       & O     & O          \\
10 & Kuantu    & B-LOC & O          \\
11 & area      & O     & B-GPE      \\
12 & of        & O     & O          \\
13 & Taipei    & B-GPE & O          \\
14 & City      & I-GPE & B-GPE      \\
15  &   -        &   -    & I-GPE     
\end{tabular}
\caption{Example of a sentence and its tagging label from \ontonotes. The prediction is generated from training a sequence-to-sequence model with a raw \bio format. }
\label{tab:seqtoseq_bio}
\end{center} 
\end{table}

\begin{table*}[]
\begin{center}
\begin{tabular}{l  r r r r r r r r }
{Statistics\textbackslash{}Dataset}   & {SNIPS} & {ATIS}  & {Ontonotes} & {CONLL} &  \\
\hline
{No. Training Samples}                & {13084} & {4478}  & {59924}     & {14041} &  \\
{No. Validation Samples}              & {700}   & {500}   & {8528}      & {3250}  &  \\
{No. Test Samples}                    & {700}   & {893}   & {8262}      & {3453}  &  \\
{Average sentence length}             & {9.05}  & {11.28} & {18.11}     & {14.53} &  \\
{\# of slot types (w/o \bio prefixes)} & {39}    & {83}    & {18}        & {4}     &  \\
{\# intent types}                     & {7}     & {21}    & {N/A}       & {N/A}   & 
\end{tabular}
\caption{Dataset Statistics} \label{tab:dataset_statistics}
\end{center}
\end{table*}

\begin{table}[]
\begin{center}
\begin{tabular}{r p{3cm}}
CONLL-2003 slot types & Natural-word label \\ \hline
LOC                   & location           \\
MISC                  & miscellaneous      \\
ORG                   & organization       \\
PER                   & person            
\end{tabular}
\caption{Label mapping for \conll{}-2003.} \label{tab:conll_labelmap}
\end{center}
\end{table}

\subsection{Shortened Generative Format} \label{supp:example_f3}
We show a failure case for a shorted generative format discussed in \ref{sec:model} where we repeat only the tagged pattern . Consider the following input $s$ and label $y$, 

\begin{table}[] 
\small
\begin{tabular}{c | c c c c c c c c c}
$s$ & These & two & men & have & two & dollars \\
\hline
$y$ & O & O & O & O & B-money & O 
\end{tabular}
\caption{Top row: original sentence. Bottom row: slot labels.} \label{tab:sample_sentence_f3}
\end{table}
If we repeat only the tagged pattern, then the output $s_g$ will be
\begin{align*}
\text{
{\small
[ \textbf{two} \textvertline \ { \color{blue} money } ]
}.
}
\end{align*}
Given $s_g$ and $s$, it is ambiguous whether the canonical label should associate with {\tt two } for {\tt two dollars} or {\tt two men}.

\begin{table*}[]
\begin{center}
\begin{tabular}{l l l p{7cm} l }
Intent/Domain              & \# sen. & \# sl. & \multicolumn{1}{c}{Slot types (without \bio tags)}                                                                                                                                                                                 &  \\ \hline
GetWeather (We)          & 2100         & 9             & timeRange, condition\_description, country, geographic\_poi, city, state, current\_location, condition\_temperature, spatial\_relation                                                      &  \\
PlayMusic (Mu)           & 2100         & 9             & genre, year, album, music\_item, playlist, service, sort, artist, track                                                                                                                     &  \\
AddToPlaylist (Pl)       & 2042         & 5             & entity\_name, music\_item, playlist, playlist\_owner, artist                                                                                                                                        &  \\
RateBook (Bo)             & 2056         & 7             & object\_type, object\_part\_of\_series\_type, object\_select, rating\_value, object\_name, best\_rating, rating\_unit                                                                           &  \\
SearchScreeningEvent (Se) & 2059         & 7             & timeRange, object\_location\_type, object\_type, location\_name, movie\_name, spatial\_relation, movie\_type                                                                                    &  \\
BookRestaurant  (Re)     & 2073         & 14            & party\_size\_number, served\_dish, timeRange, country, poi, cuisine, spatial\_relation, city, restaurant\_name, sort, restaurant\_type, facility, party\_size\_description, state &  \\
SearchCreativeWork (Cr) & 2054         & 2             & object\_type, object\_name                                                                                                                                                                                &  \\ \hline
Sum                  & 14484        & 53            &                                                                                                                                                                                                               &  \\
\# Distinct slots    &              & 39            &                                                                                                                                                                                                               & 
\end{tabular}
\caption{The intent classes and the corresponding slot labels for the \snips dataset. \# sen is the number of sentences in the entire dataset. \# sl. is the number of slot types for each  intent, excluding 'O' (no tag). Note that sentences among different intent classes can share slot types and the number of slot types in total is 39 disregarding the \bio prefixes B- and I-.} \label{tab:snips_stats}
\end{center}
\end{table*}

\begin{table*}[]
\begin{center}
\tiny
\begin{tabular}{p{2cm} p{0.3cm} p{0.3cm} p{11cm} l }
Intent              & \# sen. & \# sl. & Slot types (without \bio tags) &  \\ \hline
atis\_flight & 4298 & 71 & fromloc.city\_name, toloc.city\_name, round\_trip, arrive\_date.month\_name, arrive\_date.day\_number, stoploc.city\_name, arrive\_time.time\_relative, arrive\_time.time, meal\_description, depart\_date.month\_name, depart\_date.day\_number, airline\_name, depart\_time.period\_of\_day, depart\_date.day\_name, toloc.state\_name, depart\_time.time\_relative, depart\_time.time, depart\_date.date\_relative, or, class\_type, fromloc.airport\_name, flight\_mod, meal, economy, city\_name, airline\_code, depart\_date.today\_relative, flight\_stop, toloc.state\_code, fromloc.state\_name, toloc.airport\_name, connect, arrive\_date.day\_name, fromloc.state\_code, arrive\_date.today\_relative, depart\_date.year, depart\_time.start\_time, depart\_time.end\_time, arrive\_time.start\_time, arrive\_time.end\_time, cost\_relative, flight\_days, mod, airport\_name, aircraft\_code, toloc.country\_name, toloc.airport\_code, return\_date.date\_relative, flight\_number, fromloc.airport\_code, arrive\_time.period\_of\_day, depart\_time.period\_mod, flight\_time, return\_date.day\_name, fare\_amount, arrive\_date.date\_relative, arrive\_time.period\_mod, period\_of\_day, stoploc.state\_code, fare\_basis\_code, stoploc.airport\_name, return\_time.period\_mod, return\_time.period\_of\_day, return\_date.today\_relative, return\_date.month\_name, return\_date.day\_number, compartment, day\_name, airport\_code, stoploc.airport\_code, flight \\ 
atis\_airfare & 471 & 45 & round\_trip, fromloc.city\_name, toloc.city\_name, cost\_relative, fare\_amount, class\_type, economy, airline\_name, flight\_mod, depart\_time.time\_relative, depart\_time.time, arrive\_date.month\_name, arrive\_date.day\_number, airline\_code, flight\_number, stoploc.city\_name, toloc.airport\_name, depart\_date.date\_relative, depart\_date.day\_name, depart\_date.month\_name, depart\_date.day\_number, toloc.state\_code, depart\_time.period\_of\_day, flight\_stop, fromloc.state\_name, toloc.state\_name, toloc.airport\_code, aircraft\_code, depart\_date.year, arrive\_time.time\_relative, arrive\_time.time, fromloc.airport\_code, fromloc.airport\_name, depart\_date.today\_relative, return\_date.month\_name, return\_date.day\_number, connect, meal,  arrive\_date.date\_relative, arrive\_date.day\_name, or, depart\_time.period\_mod, flight\_time, flight\_days, fromloc.state\_code \\ 
atis\_ground\_service & 291 & 23 & toloc.airport\_name, city\_name, fromloc.airport\_name, toloc.city\_name, state\_code, transport\_type, airport\_name, fromloc.city\_name, or, depart\_date.date\_relative, depart\_date.day\_name, time, depart\_date.month\_name, depart\_date.day\_number, today\_relative, flight\_time, state\_name, period\_of\_day, time\_relative, day\_name, month\_name, day\_number, airport\_code \\ 
atis\_airline & 195 & 36 & fromloc.city\_name, toloc.city\_name, toloc.state\_code, mod, depart\_date.day\_name, class\_type, depart\_date.today\_relative, stoploc.city\_name, aircraft\_code, arrive\_date.month\_name, arrive\_date.day\_number, toloc.airport\_name, fromloc.state\_code, depart\_time.period\_of\_day, airline\_code, flight\_number, depart\_time.time\_relative, depart\_time.time, depart\_date.month\_name, depart\_date.day\_number, arrive\_time.time, city\_name, airport\_name, flight\_stop, arrive\_time.period\_of\_day, fromloc.airport\_code, airline\_name, depart\_date.date\_relative, connect, flight\_days, round\_trip, cost\_relative, fromloc.airport\_name, depart\_time.start\_time, depart\_time.end\_time, toloc.state\_name \\ 
atis\_abbreviation & 180 & 14 & fare\_basis\_code, airport\_code, airline\_code, meal, meal\_code, restriction\_code, airline\_name, aircraft\_code, class\_type, days\_code, mod, fromloc.city\_name, toloc.city\_name, booking\_class \\ 
atis\_aircraft & 90 & 23 & fromloc.city\_name, toloc.city\_name, depart\_time.time\_relative, depart\_time.time, toloc.state\_code, airline\_name, mod, class\_type, depart\_date.day\_name, airline\_code, flight\_number, stoploc.city\_name, depart\_time.period\_of\_day, flight\_mod, aircraft\_code, arrive\_time.time\_relative, arrive\_time.time, arrive\_date.day\_name, depart\_date.month\_name, depart\_date.day\_number, arrive\_date.month\_name, arrive\_date.day\_number, city\_name \\ 
atis\_flight\_time & 55 & 20 & flight\_time, airline\_name, toloc.airport\_code, depart\_date.month\_name, depart\_date.day\_number, fromloc.city\_name, toloc.city\_name, depart\_date.day\_name, depart\_time.period\_of\_day, airline\_code, flight\_number, flight\_mod, depart\_date.date\_relative, depart\_time.time, fromloc.airport\_name, aircraft\_code, depart\_time.time\_relative, airport\_name, class\_type, meal\_description \\ 
atis\_quantity & 54 & 25 & airline\_code, class\_type, flight\_stop, fromloc.city\_name, toloc.city\_name, arrive\_date.month\_name, arrive\_date.day\_number, depart\_date.month\_name, depart\_date.day\_number, economy, airline\_name, round\_trip, toloc.airport\_name, depart\_date.today\_relative, arrive\_time.time\_relative, arrive\_time.time, fare\_basis\_code, city\_name, stoploc.city\_name, flight\_number, flight\_days, depart\_time.time\_relative, depart\_time.time, depart\_time.period\_of\_day, aircraft\_code \\ 
atis\_airport & 38 & 9 & city\_name, state\_code, fromloc.city\_name, mod, airport\_name, toloc.city\_name, flight\_stop, state\_name, airline\_name \\ 
atis\_capacity & 37 & 5 & fromloc.city\_name, toloc.city\_name, airline\_name, aircraft\_code, mod \\ 
atis\_flight, atis\_airfare & 33 & 21 & fromloc.city\_name, toloc.city\_name, airline\_name, flight\_number, depart\_date.day\_name, depart\_date.month\_name, depart\_date.day\_number, flight\_mod, round\_trip, depart\_time.time\_relative, depart\_time.time, cost\_relative, fare\_amount, depart\_date.date\_relative, arrive\_time.time\_relative, arrive\_time.time, depart\_time.period\_of\_day, toloc.state\_code, flight\_stop, return\_date.date\_relative, return\_date.day\_name \\ 
atis\_distance & 30 & 8 & fromloc.airport\_name, toloc.city\_name, fromloc.city\_name, depart\_time.time, depart\_date.month\_name, depart\_date.day\_number, city\_name, airport\_name \\ 
atis\_city & 25 & 11 & city\_name, airline\_name, airport\_code, fromloc.airport\_code, fromloc.city\_name, toloc.city\_name, depart\_time.time\_relative, depart\_time.time, depart\_time.period\_of\_day, airport\_name, class\_type \\ 
atis\_ground\_fare & 25 & 6 & transport\_type, city\_name, fromloc.city\_name, fromloc.airport\_name, airport\_name, toloc.city\_name \\ 
atis\_flight\_no & 20 & 22 & toloc.city\_name, fromloc.city\_name, arrive\_time.time\_relative, arrive\_time.time, fromloc.state\_name, toloc.state\_name, depart\_date.day\_name, depart\_date.month\_name, depart\_date.day\_number, airline\_name, depart\_time.time, flight\_mod, toloc.state\_code, flight\_time, cost\_relative, class\_type, depart\_time.time\_relative, depart\_time.period\_of\_day, stoploc.city\_name, flight\_number, or, depart\_date.today\_relative \\ 
atis\_meal & 12 & 12 & meal, fromloc.city\_name, toloc.airport\_code, airline\_name, flight\_number, toloc.city\_name, airline\_code, arrive\_time.time, toloc.state\_code, depart\_date.day\_name, depart\_time.period\_of\_day, meal\_description \\ 
atis\_restriction & 6 & 6 & restriction\_code, cost\_relative, round\_trip, fromloc.city\_name, toloc.city\_name, fare\_amount \\ 
atis\_airline, atis\_flight\_no & 2 & 7 & fromloc.city\_name, toloc.city\_name, depart\_date.date\_relative, depart\_date.month\_name, depart\_date.day\_number, arrive\_time.time\_relative, arrive\_time.time \\ 
atis\_day\_name & 2 & 2 & fromloc.city\_name, toloc.city\_name \\ 
atis\_aircraft, atis\_flight,atis\_flight\_no & 1 & 5 & fromloc.city\_name, toloc.city\_name, airline\_name, depart\_time.time\_relative, depart\_time.time \\ 
atis\_cheapest & 1 & 1 & cost\_relative \\ 
atis\_ground\_service, atis\_ground\_fare & 1 & 1 & fromloc.airport\_name \\ 
atis\_airfare, atis\_flight\_time & 1 & 3 & flight\_time, fromloc.city\_name, toloc.city\_name \\ 
atis\_airfare, atis\_flight & 1 & 4 & airline\_name, flight\_number, fromloc.airport\_code, toloc.airport\_code \\ 
atis\_flight, atis\_airline & 1 & 5 & fromloc.city\_name, toloc.city\_name, depart\_date.day\_name, depart\_time.time\_relative, depart\_time.time \\ 
atis\_flight\_no, atis\_airline & 1 & 5 & fromloc.city\_name, toloc.city\_name, depart\_date.day\_name, depart\_time.time\_relative, depart\_time.time \\ 
                                       &  \\ \hline
Sum                  & 5871        & 390            &                                                                                                                                                                                                               &  \\
\# Distinct slots    &              & 83            &                                                                                                                                                                                                               & 
\end{tabular}
\caption{The intent classes and the corresponding slot labels for the \atis dataset. \# sen is the number of sentences in the entire dataset. \# sl. is the number of slot types for each  intent, excluding `O' (no tag). Note that sentences among different intent classes can share slot types and the number of slot types in total is 83 disregarding the \bio prefixes B- and I-.} \label{tab:atis_stats}
\end{center}
\end{table*}

\subsection{Dataset Details} \label{sec:supp_data}
We provide details on the statistics of datasets used in Table \ref{tab:dataset_statistics}. We provide details on the intent and slot label types in Table \ref{tab:snips_stats} for the \snips dataset and Table \ref{tab:atis_stats} for the \atis dataset. 
The tagging label types and their label mapping are listed in Table \ref{tab:ontonotes_labelmap} for \ontonotes and \ref{tab:conll_labelmap} for \conll.

\begin{table*}[]
\begin{center}
\begin{tabular}{r p{4cm} p{6cm}}
Ontonotes slot types & Natural-word label                    & Descriptions                                         \\ \hline
CARDINAL             & cardinal                              & Numerals that do not fall under another type         \\
DATE                 & date                                  & Absolute or relative dates or periods                \\
EVENT                & event                                 & Named hurricanes, battles, wars, sports events, etc. \\
FAC                  & facility                              & Buildings, airports, highways, bridges, etc.         \\
GPE                  & country city state                    & countries, cities, states                            \\
LANGUAGE             & language                              & Any named language                                   \\
LAW                  & law                                   & Named documents made into laws                       \\
LOC                  & location                              & Non-GPE locations, mountain ranges, bodies of water  \\
MONEY                & money                                 & Monetary values, including unit                      \\
NORP                 & nationality religious political group & Nationalities or religious or political groups       \\
ORDINAL              & ordinal                               & "first", "second"                                    \\
ORG                  & organization                          & Companies, agencies, institutions, etc.              \\
PERCENT              & percent                               & Percentage (including ``\%'')                        \\
PERSON               & person                                & People, including fictional                          \\
PRODUCT              & product                               & Vehicles, weapons, foods, etc. (Nor services)        \\
QUANTITY             & quantity                              & Measurements, as of weight or distance               \\
TIME                 & time                                  & Times smaller than a day                             \\
WORK\_OF\_ART        & work of art                           & Titiles of books, songs, etc.                       
\end{tabular}
\caption{Label mapping for \ontonotes. The descriptions are obtained from \citet{ontonotes_release5} } 
\label{tab:ontonotes_labelmap}
\end{center}
\end{table*}

\subsection{Low-Resource Results} \label{sec:sup_low_resource}
We provide full numeric results for low-resource experiments from Section \ref{sec:limited_resource} and  \ref{sec:supervised_transfer_learning} in Table \ref{tab:snips_lowresource} and Table \ref{tab:snips_lowresource_suppretrain_ner} respectively.

\begin{table*}[]
\begin{center}
\begin{tabular}{lcccccccccccc}
      &   & &   \multicolumn{8}{c}{\textbf{Slot Labeling F1 Score }}                                                                                                                                                                                                                                     \\ \cmidrule(lr){4-11}
 \textbf{\%}   & \textbf{no. sen} & \textbf{s/t} & \textbf{\ours} & $\pm$ & \textbf{\ours-o} & $\pm$ & \textbf{\ours-n} & $\pm$ & {\sc \textbf{BERT}}  &  $\pm$   \\
0.25 & 33    & 0.84   & 60.37 & 2.66 & 57.49 & 2.37 & 50.12 & 5.29 & 44.73 & 6.43 \\
0.5  & 65    & 1.68   & 73.69 & 1.51 & 72.02 & 1.22 & 67.03 & 2.41 & 60.12 & 2.77 \\
1    & 131   & 3.35   & 83.53 & 1.63 & 83.32 & 0.68 & 81.52 & 0.60 & 73.94 & 1.34 \\
2    & 262   & 6.71   & 89.06 & 0.79 & 88.69 & 1.07 & 87.97 & 1.21 & 84.30 & 0.97 \\
5    & 654   & 16.77  & 93.35 & 0.21 & 92.98 & 0.45 & 92.47 & 0.30 & 90.99 & 0.69 \\
10   & 1308  & 33.55  & 94.65 & 0.18 & 94.51 & 0.29 & 94.41 & 0.22 & 93.56 & 0.68 \\
20   & 2617  & 67.10  & 95.55 & 0.20 & 95.14 & 0.24 & 95.55 & 0.36 & 94.77 & 0.15 \\
40   & 5234  & 134.19 & 96.15 & 0.15 & 96.22 & 0.10 & 96.16 & 0.11 & 96.05 & 0.29 \\
100  & 13084 & 335.49 & 96.71 & 0.05 & 96.96 & 0.10 & 97.01 & 0.10 & 96.73 & 0.16         
\end{tabular}
\caption{
Test results of our models for under varying levels of training resources on the \snips training data. \textbf{\%} and \textbf{no. sen} columns indicate the \% of the original training data and number of training sentences. \textbf{s/t} indicates the number of sentences per slot label type. 
} \label{tab:snips_lowresource} 
\end{center}
\end{table*}

\begin{table*}[]
\begin{center}
\begin{tabular}{rrr ccccccccc}
      &   & &   \multicolumn{8}{c}{\textbf{Slot Labeling F1 Score }}                                                                                                                                                                                                                                     \\ \cmidrule(lr){4-11}
 \textbf{\%}   & \textbf{no. sen} & \textbf{s/t} & \textbf{\ours} & $\pm$ & \textbf{\ours-o} & $\pm$ & \textbf{\ours-n} & $\pm$ & {\sc \textbf{BERT}}  &  $\pm$   \\
0.25 & 33                                         & 0.8                         & 63.83                         & 2.62                      & 62.67                                      & 2.06                      & 57.39                                         & 2.90                      & 46.27                                & 3.56                      \\
0.5  & 65                                         & 1.7                         & 76.60                         & 0.79                      & 75.89                                      & 1.32                      & 72.46                                         & 2.70                      & 60.25                                & 1.89                      \\
1    & 131                                        & 3.4                         & 84.90                         & 0.71                      & 85.08                                      & 0.20                      & 84.69                                         & 0.62                      & 73.55                                & 1.50                      \\
2    & 262                                        & 6.7                         & 89.37                         & 0.65                      & 89.24                                      & 0.32                      & 89.36                                         & 0.65                      & 82.63                                & 1.42                      \\
5    & 654                                        & 16.8                        & 93.74                         & 0.23                      & 93.21                                      & 0.17                      & 93.63                                         & 0.32                      & 89.83                                & 0.43                      \\
10   & 1308                                       & 33.5                        & 94.78                         & 0.24                      & 94.60                                      & 0.41                      & 94.78                                         & 0.23                      & 92.23                                & 0.35                      \\
20   & 2617                                       & 67.1                        & 95.70                         & 0.32                      & 95.66                                      & 0.18                      & 95.87                                         & 0.18                      & 94.39                                & 0.34                      \\
40   & 5234                                       & 134.2                       & 96.52                         & 0.25                      & 96.60                                      & 0.23                      & 96.40                                         & 0.21                      & 95.67                                & 0.32                      \\
100  & 13084                                      & 335.5                       & 97.27                         & 0.05                      & 97.16                                      & 0.18                      & 97.29                                         & 0.20                      & 96.44                                & 0.23                     
\end{tabular}
\caption{Test results of our models for under varying levels of training resources on the \snips training data, with supervised pre-training on \ontonotes NER dataset. \textbf{\%} and \textbf{no. sen} columns indicate the \% of the original training data and number of training sentences. \textbf{s/t} indicates the number of sentences per slot label type. 
 } \label{tab:snips_lowresource_suppretrain_ner}
\end{center}
\end{table*}

\subsection{Few-Shot Experiments} \label{sec:fs_full_results}
We provide details on the data used for our few-shot experiments and the full results in this Section. 

\subsubsection{Episode Data} \label{sec:fs_episode}

We use two data constructions: the original episodes from \citet{fewshot_sl_hou} and our own constructed episodes with \citet{fewshot_sl_hou}'s definition. We provide the episode statistics (Ave $|\S|$) in Table \ref{tab:five_shot_all} for both constructions which demonstrate that the support size are comparable for each domain. The major difference is that \citet{fewshot_sl_hou} pre-processes data by lowercasing all letters and removing extra tokens such as commas and apostrophes. In addition, \citet{fewshot_sl_hou} modify the BIO prefixes in cases where the tokenization splits a token with the ``B-'' prefix into two or more units. For instance, the token [``lora's''] with tag [\textit{B-playlist\_owner}] becomes [``lora'', ``s''] with tags [\textit{B-playlist\_owner}, \textit{I-playlist\_owner}]. This treatment considerably increases the number of tokens with ``I-" tags in the episodes created by \citet{fewshot_sl_hou}. 
Both data have $100$ episodes with $20$ query sentences. 
We provide the results for our episodes with lowercased words and Hou's episodes, which shows the similarity between two settings.

We note that the support size for some domains can be smaller than in other domains, according to \citet{fewshot_sl_hou} K-shot definition. For instance, domain {\bf \sc Cr} has around $5$ sentences on average whereas domain {\bf \sc Re} has more than $30$ sentences. This is because for some domains, there can be many tags of the same types in a single sentence.

\subsubsection{5-Shot Results}  \label{sec:five_shot} 
Table \ref{tab:five_shot_all} details the full results on both episode data and multiple variations of our models. \textbf{\ours + \snips} is trained on leave-one-out dataset. For instance, when we evaluate on domain \textbf{We}, we train on other domains except for \textbf{We}. This is the setting used in all other baselines. \textbf{\ours + Onto} shows the results trained on \ontonotes NER task (see label types in Table \ref{tab:snips_stats}), which is from a different domain that the \snips dataset. \textbf{\ours w/o meta} involves no additional meta-training and fine-tunes on our backbone model directly.

\begin{table*}[]
\begin{center}
\begin{tabular}{p{0.22cm} lcccccccccccc}
&        & \textbf{We} & \textbf{Mu} & \textbf{Pl} & \textbf{Bo} & \textbf{Se} & \textbf{Re} & \textbf{Cr} & \textbf{Ave.} \\ \cmidrule(lr){2-10}
\multirow{12}{*}{ \rotatebox{90}{ \citet{fewshot_sl_hou}'s Episodes} } 
& Ave. $|\S|$   &      28.91                &  34.43                      &   13.84                     &   19.83                     &   19.27                     &   41.58                     &    5.28                   &                          \\ 
& SimBERT      & 53.46                  & 54.13                  & 42.81                  & 75.54                  & 57.10                  & 55.30                  & 32.38                  & 52.96                    \\
& TransferBERT & 59.41                  & 42.00                  & 46.70                  & 20.74                  & 28.20                  & 67.75                  & 58.61                  & 46.11                    \\
& Matching Net & 36.67                  & 33.67                  & 52.60                  & 60.09                  & 38.42                  & 33.28                  & 72.10                  & 47.98                    \\
& WPZ + BERT   & 67.82                  & 55.99                  & 46.02                  & 72.17                  & 73.59                  & 60.18                  & 66.89                  & 63.24                    \\
& L-TapNet+CDT & \textbf{71.64}      & \textbf{67.16}      & \textbf{75.88}   & \textbf{84.38}   & \textbf{82.58}   & \textbf{70.05}   & \textbf{73.41}                  & \textbf{75.01}                    \\ \cmidrule(lr){2-10}
& \ours+ \snips & {\ul 87.66}                & {\ul 81.62}                & {\ul 83.37}                & {\ul 89.72}                & {\ul 86.80}                & {\ul 86.14}                &  73.02   & {\ul 84.05}             \\
 &       $\pm$    & 7.35                 & 9.75                 & 8.75                 & 5.23                 & 8.33                 & 6.30                 & 11.79                \\
& \ours+ Onto &     {\ul 77.87} &     {\ul 79.71} &     {\ul 79.87} &     84.18 &     75.84 &     {\ul 78.71} &     40.23 & 73.77 \\
 &       $\pm$ &      7.98 &      8.54 &      9.08 &      5.64 &      8.35 &      7.54 &     14.59 &    \\
 &  \ours + No Meta &     70.82 &     {\ul 74.24} &     73.88 &     83.32 &     74.94 &     {\ul 75.18} &     41.22 & 70.51 \\
 &       $\pm$ &      8.39 &      9.26 &      8.11 &      6.11 &      9.93 &      7.24 &     16.42 &    \\
%\cmidrule(lr){2-10}
[0.4cm] 
\multirow{13}{*}{ \rotatebox{90}{Our episodes} } 
&  Ave. $|\S|$  &      24.92                &  32.49                      &   12.56                     &   18.44                     &   17.18                     &   37.06                     &    5.70                   &                          \\ 
& \ours + \snips &{\ul \textbf{91.35}} & {\ul \textbf{86.73}} & {\ul \textbf{87.20}} & {\ul \textbf{95.85}} & {\ul \textbf{92.71}} & {\ul \textbf{91.23}} & {\ul \textbf{91.55}} & \multicolumn{1}{r}{{\ul \textbf{90.95}}} \\
& $\pm$                      & 6.13                 & 6.52                 & 6.59                 & 3.13                 & 5.45                 & 5.86                 & 7.25                 &                                         \\
& \ours + Onto  & {\ul 83.15} & {\ul {86.15}} & {\ul {80.36}} & {\ul {90.27}} & {\ul {84.87}} & {\ul {85.89}} & 68.08 & {\ul {82.68}} \\
& $\pm$   & 6.52                 & 6.15                 & 10.87                & 5.12                 & 7.98                 & 6.52                 & 24.50 &                      \\
& \ours + No Meta   & {\ul 73.14}          & {\ul 82.02}          & {\ul 78.82}          & {\ul 84.86}          & {\ul 83.14}          & {\ul 86.63}          & 52.56 & {\ul 77.31}          \\
& $\pm$   & 9.41                 & 5.09                 & 9.94                 & 3.55                 & 4.87                 & 8.52                 & 22.06 &  \\
& \ours-l + \snips  	& {\ul 86.18}                & {\ul 83.51}                & {\ul 84.15}                & {\ul 89.33}                & {\ul 85.74}                & {\ul 86.34}                & {\ul 76.69}                & {\ul 84.60} \\
& $\pm$    	  	& 6.76                 & 7.63                 & 6.67                 & 5.31                 & 7.46                 & 6.53                 & 11.34                &                          \\
& \ours-l + Onto &     {\ul 76.07} &     {\ul 79.25} &     {\ul 77.54} &     82.49 &     74.92 &     {\ul 80.44} &     42.78 & 73.36 \\
& $\pm$ &      9.21 &     10.10 &      9.19 &      6.45 &     10.17 &      6.37 &     16.56 &    \\
& \ours-l + No Meta &     68.56 &     {\ul 76.17} &     74.63 &     81.02 &     74.47 &    {\ul  78.88} &     38.60 & 70.33 \\
& $\pm$ &     11.06 &      9.20 &      8.21 &      7.41 &      9.16 &      7.67 &     17.39 &    \\
\end{tabular}
\caption{Our 5-shot slot tagging results on $7$ domains of the \snips dataset. We provide the average and the standard deviation of F1 scores over 100 episodes. \ours-l indicate that we use lowercased words for input sentences. 
} \label{tab:five_shot_all}
\end{center} 
\end{table*}

\subsection{1-Shot Results} \label{sec:one_shot}

Our models tend to perform well when there are sufficient enough sentences to fine-tune on. For domain `Cr' (SearchCreativeWork) where there are highly limited number of sentences ($5$ sentences on average), our model does not perform well compared to other baselines. 
This observation is consistent with the 1-shot results, which we include in the appendix Section \ref{sec:one_shot} Table \ref{tab:one_shot}, where we typically have less than $10$ sentences in the support set. In this case, our model performs comparable to Warm Proto Zero with \bert on average but is outperformed by L-TapNet+CDT.
Other techniques to improve on the 1-shot learning result include bootstrapping more sentences from unlabeled corpus with the labels from the support set for better optimization.

\begin{table*}[]
\begin{center}
\begin{tabular}{p{0.22cm} lcccccccccccc}
&        & \textbf{We} & \textbf{Mu} & \textbf{Pl} & \textbf{Bo} & \textbf{Se} & \textbf{Re} & \textbf{Cr} & \textbf{Ave.} \\ \cmidrule(lr){2-10}
\multirow{12}{*}{ \rotatebox{90}{ \citet{fewshot_sl_hou}'s Episodes} } 
 & Ave. $|\S|$     & 6.15                        & 7.66                      & 2.96                              & 4.34                         & 4.29                                     & 9.41                               & 1.30                                   &                                    \\
                              & SimBERT            & 36.10                       & 37.08                     & 35.11                             & 68.09                        & 41.61                                    & 42.82                              & 23.91                                  & \multicolumn{1}{r}{40.67}          \\
                              & TransferBERT       & 55.82                       & 38.01                     & 45.65                             & 31.63                        & 21.96                                    & 41.79                              & 38.53                                  & \multicolumn{1}{r}{39.06}          \\
                              & Matching Net       & 21.74                       & 10.68                     & 39.71                             & 58.15                        & 24.21                                    & 32.88                              & \textbf{69.66}                         & \multicolumn{1}{r}{36.72}          \\
                              & WPZ + BERT         & 46.72                       & 40.07                     & 50.78                             & 68.73                        & 60.81                                    & 55.58                              & 67.67                                  & \multicolumn{1}{r}{55.77}          \\
                              & L-TapNet+CDT       & \textbf{71.53}              & \textbf{60.56}            & \textbf{66.27}                    & \textbf{84.54}               & \textbf{76.27}                           & \textbf{70.79}                     & 62.89                                  & \multicolumn{1}{r}{\textbf{70.41}} \\
\cmidrule(lr){2-10}                              
& \ours+ \snips  & {\ul 77.81}                & {\ul 74.66}                & {\ul 67.81}                & 79.60                & 71.05                & {\ul 78.61}                & 64.07                & {\ul 73.37} \\
& $\pm$        & 7.21                 & 7.46                 & 7.22                 & 6.13                 & 7.88                 & 5.58                 & 10.53                &                           \\
& \ours+ Onto   &     50.56 &     {\ul 68.54} &     56.11 &     70.17 &     55.66 &     64.78 &     12.86 & 54.10 \\
  &      $\pm$ &      9.53 &      6.75 &     10.52 &      5.96 &      5.76 &      6.65 &     11.45 &    \\
& \ours + No Meta &     40.28 &     53.42 &     45.71 &     63.90 &     42.57 &     59.29 &     14.11 & 45.61 \\
& $\pm$  &      9.23 &     10.71 &     11.93 &      8.26 &      7.94 &      7.45 &     11.50 &    \\
%\cmidrule(lr){2-10}
[0.4cm] 
\multirow{13}{*}{ \rotatebox{90}{Our episodes} } 
&  Ave. $|\S|$  & 6.17. & 6.99 & 2.66 & 3.73 & 4.15 & 9.07 & 1.27 \\
                              & \ours + SNIPS       & \textbf{{\ul 82.62}} & \textbf{{\ul 77.46}} & \textbf{{\ul 71.33}} & \textbf{{\ul 85.49}} & \textbf{{\ul 83.22}} & \textbf{{\ul 84.23}} & \textbf{{\ul 82.92}} & \textbf{{\ul 81.04}} \\
                              &  $\pm$                   & 5.96  & 7.56  & 7.15  & 5.74  & 6.47  & 5.67  & 8.71  &                           \\
                              & \ours + Onto        & 56.39 & {\ul 67.10} & 53.49 & 71.94 & 66.21 & 69.04 & 28.80 & 59.00 \\
                              & $\pm$                   & 10.67 & 7.50  & 11.18 & 7.43  & 10.02 & 7.68  & 17.67 &                           \\
                              & \ours  + No Meta          & 46.42 & 59.02 & 47.47 & 63.79 & 49.42 & 64.45 & 17.60 & 49.74 \\
                              &  $\pm$                   & 9.52  & 9.45  & 11.01 & 7.33  & 10.32 & 7.63  & 7.63  &                           \\
                              & \ours-l + SNIPS  & {\ul 77.42} & {\ul 73.45} & {\ul 67.05} & 76.85 & 72.54 & {\ul 79.54} & 63.44 & {\ul 72.90} \\
                              &  $\pm$                   & 6.09  & 8.18  & 8.26  & 5.93  & 8.97  & 5.62  & 10.19 &    \\
                              & \ours-l + Onto  &     50.65 &     {\ul 62.58} &     50.55 &     65.42 &     50.23 &     61.82 &     18.59 & 51.41 \\
   			     &     $\pm$ &      8.47 &      7.94 &      8.32 &      6.71 &      9.83 &      6.92 &     12.15 &    \\
                              & \ours-l + No Meta &     41.47 &     57.69 &     44.13 &     61.22 &     40.96 &     59.60 &     11.00 & 45.15 \\
   			     &     $\pm$ &      8.25 &     12.46 &      7.45 &      6.55 &     11.50 &      8.60 &     10.87 &    \\
\end{tabular}
\caption{Our 1-shot slot tagging results on \citet{fewshot_sl_hou}'episodes and our own constructed episodes. We provide the average and the standard deviation of F1 scores over 100 episodes.  \ours-l indicate that we use lowercased words for input sentences. 
} \label{tab:one_shot}
\end{center} 
\end{table*}

\begin{comment}
\subsection{Zero-Shot Examples} \label{sec:zero_shot}

\begin{table}[h]
\begin{center}
\begin{tabular}{c c c }
\textbf{Words}      & \textbf{Label}    & \textbf{Prediction}    \\
a          & O        & O             \\
movie      & O        & O             \\
starring   & O        & O             \\
britney    & B-actor  & B-person      \\
spears     & I-person & I-person      \\
\hline
when       & O        & O             \\
did        & O        & O             \\
underworld & B-title  & B-movie name  \\
come       & O        & O             \\
out        & O        & O             \\
\hline
show       & O        & O             \\
me         & O        & O             \\
classic    & B-genre  & B-object type \\
comedies   & I-genre  & I-object type \\
starring   & O        & O             \\
bill       & B-actor  & B-object name \\
murray     & I-actor  & I-object type
\end{tabular}
\caption{Examples of predictions on sentences from MIT Movie corpus, using models trained on \snips, \atis, \conll, and \ontonotes. Our model exhibit generalization performance where the predictions closely match the expected labels, albeit with different label type convention. 
}
\end{center}
\end{table}
\end{comment}

\end{document}